# BERT-Deep CNN: State-of-the-Art for Sentiment Analysis of COVID-19 Tweets

**Javad Hassannataj Joloudari[1], Sadiq Hussain[2], Mohammad Ali Nematollahi[3], Rouhollah Bagheri[4,*], Fatemeh Fazl[5], Roohallah Alizadehsani[6], Reza Lashgari[7], Ashis Talukder[8*]**

[1]Department of Computer Engineering, Faculty of Engineering, University of Birjand, Birjand 9717434765, Iran
[2]Dibrugarh University, Assam 786004, India;
[3]Department of Computer Sciences, Fasa University, Fasa, Iran
[4]Department of Management, Faculty of Economics and Administrative Sciences, Ferdowsi University of Mashhad
[5]Department of Electronic Engineering, Faculty of Electrical and Computer Engineering, University of Birjand, Birjand, Iran
[6] Institute for Intelligent Systems Research and Innovation, Deakin University, Geelong, VIC, Australia
[7]Institute of Medical Science and Technology, Shahid Beheshti University, Tehran, Iran
[8]Statistics Discipline, Khulna University, Khulna-9208, Bangladesh

*** Correspondence:**
ashistalukder3168@stat.ku.ac.bd; rbagheri@um.ac.ir

## Abstract
The free flow of information has been accelerated by the rapid development of social media technology. There has been a significant social and psychological impact on the population due to the outbreak of Coronavirus disease (COVID-19). The COVID-19 pandemic is one of the current events being discussed on social media platforms. In order to safeguard societies from this pandemic, studying people's emotions on social media is crucial. As a result of their particular characteristics, sentiment analysis of texts like tweets remains challenging. Sentiment analysis is a powerful text analysis tool. It automatically detects and analyzes opinions and emotions from unstructured data. Texts from a wide range of sources are examined by a sentiment analysis tool, which extracts meaning from them, including emails, surveys, reviews, social media posts, and web articles. To evaluate sentiments, natural language processing (NLP) and machine learning techniques are used, which assign weights to entities, topics, themes, and categories in sentences or phrases. Machine learning tools learn how to detect sentiment without human intervention by examining examples of emotions in text. In a pandemic situation, analyzing social media texts to uncover sentimental trends can be very helpful in gaining a better understanding of society's needs and predicting future trends. We intend to study society's perception of the COVID-19 pandemic through social media using state-of-the-art BERT and Deep CNN models. The superiority of BERT models over other deep models in sentiment analysis is evident and can be concluded from the comparison of the various research studies mentioned in this article.

## Introduction

The coronavirus disease 2019 (COVID-19) has been a calamitous event with a prime impact on all spheres of life, including the economy, mental health, and society [1]. The abrupt socio-economic changes motivated researchers across the globe to gain insights into various aspects of the COVID-19 pandemic. The potential of machine learning has been prominent during this period [2]. Also, deep learning models have played a crucial role in predicting the number of active cases and deaths [3].

Besides, sentiment analysis utilizes text analytics and natural language processing (NLP) tools to extract and identify the writers' feelings from their tweets or messages [4]. These sentiments can be neutral, positive, or negative. Social media, including Twitter, has been a tool to express the public's feelings during the pandemic. Social media tweets and posts brought another level of understanding when integrated with sentiment analysis [5]. During the early phase of the pandemic, sentiments such as anxiety and fear were dominant. Twitter-based sentiment analysis exploited topic modeling. Region-specific studies include sentiment analysis in Nepal [6] and Australia [7], which demonstrated positive sentiments with an element of fear. In another study, sentiment analysis reviewed the impact of the digital platform during COVID-19 in Spain [8]. Another analysis confirmed that the mood deterioration correlated with the announcement of the lockdown that recovered within a brief period [9].

Social media platforms such as Twitter, Instagram, Reddit, and Facebook have been immensely indulged in examining and fast-checking to combat misinformation due to the emergence of bizarre conspiracy theories [10]. The analytics tools to sojourn such misinformation were the need of the hour in pandemic scenarios. The temperament and the prevailing mood of the general human population could be revealed by analyzing social media space. In the NLP domain, pre-trained language models such as BERT (Bidirectional Encoder Representations from Transformers) showcase their efficacy [11]. These language models are pre-trained on a considerable amount of unannotated text and yield superior performance with a mitigated requirement of labeled data, and provide faster training than conventional training. BERT has recorded state-of-the-art performance in NLP tasks, including sentiment analysis, by substantial margins [12].

Procedures for overall pre-training and fine-tuning BERT are shown in Figure 1 [13]. The same architectures are utilized for pre-training and fine-tuning, except for output layers [14-16]. Models are initialized for various downstream tasks using the same pre-trained model parameters. All parameters are adjusted during fine-tuning. Every input example now has the special symbol [CLS] before it, and [SEP] is a unique token that separates questions and answers [13].

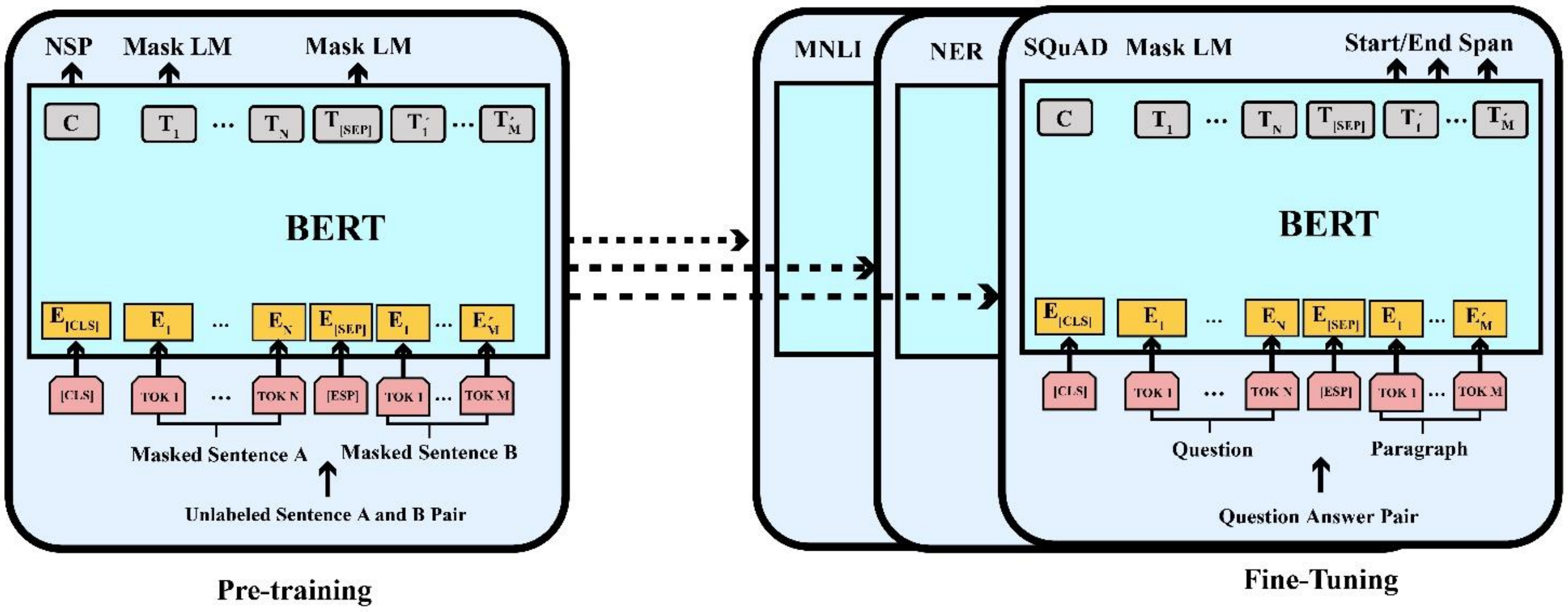

Figure 1. Procedures for overall pre-training and fine-tuning BERT.

People with COVID-19 symptoms must get examined and isolate themselves to mitigate the spread [17, 18]. Social media is a way to express their feelings during the isolation period. Data from social media may be misleading at times, although it contains valuable and real-time information as well [19]. Misleading information may lead to a new height in sufferings for such patients. A COVID-19 patient passed through several mental and physical traumas. The need for utilizing logical strategies to combat misinformation is high. Online social media platforms such as Twitter contain various noisy data. After cleaning the data, it is found that data capture human emotions, feelings, expressions, and thoughts. Twitter-data-based sentiment analysis can explore public sentiments to examine the fear associated with the disease. This motivates us to review the state-of-the-art BERT and Deep CNN-based sentiment analysis of COVID-19-based tweets realizing the significance of such research.
The main contribution of the study is as follows:

1. We reviewed the state-of-the-art BERT and deep CNN models for sentiment analysis of COVID-19 tweets.
2. Future research directions for devising a lightweight, high-quality BERT model are proposed.

In the following, the advantages and limitations of the studies for sentiment analysis of COVID-19 Tweets have been described in Table 1.

**Table1.** Papers related to COVID-19-based Sentiment Analysis.

| Ref | Year | Topic of discussion | Advantages/Limitations | Performance |
|---|---|---|---|---|
| D'Andrea et al. [20] | 2019 | Sensing the public opinion on vaccination automatically from tweets | Bag-of-words and SVM for classification were utilized | Accuracy: 65.4% |
| Zhang et al. [21] | 2020 | The exploitation of the largest Twitter English depression dataset | Their models could be readily applied to the monitoring of stress and depression trends in | Accuracy 76.5% (BERT) |
| | | | targeted groups. XLNet, RoBERTa, and BERT were used. | |
| Chatsiou [22] | 2020 | Auto-assign sentences for the corpus of the COVID-19 press briefing | CNN classifiers integrated with transformers like BERT outperformed models with other embeddings (Word2Vec, Glove, ELMo) | Accuracy 68.65% (CNN+BERT) |
| Jelodar, H. et al. [23] | 2020 | Based on COVID-19 comments, deep sentiment classification was performed | LSTM achieved superior classification compared to its machine-learning counterparts. | Accuracy 81.15% |
| Chakraborty et al. [24] | 2020 | Sentiment analysis based on 226668 tweets related to COVID-19 | Employing a fuzzy rule base for Gaussian membership | Accuracy 81% |
| Kairon et al. [25] | 2021 | Comparison of continuous-variable quantum neural Networks and quantum backpropagation multilayer perceptron (QBMLP) | Significant results were demonstrated on complex and sporadic data. | P-value 0.9849 for QBMLP Model on Indian confirmed case prediction |
| Abdelminaam et al. [26] | 2021 | Fake news on the COVID-19 identification system | Modified GRU and Modified-LSTM were utilized to enhance the accuracy | Accuracy 98.6% (LSTM two layers) |
| Garcia et al. [27] | 2021 | Positive and negative emotions of COVID-19 were explored. | They integrated recent embedding models (SBERT, mUSE, Fast-Text) for feature extraction. | Accuracy 83% |
| Naseem et al. [10] | 2021 | They developed a novel massive sentiment data set, COVIDSENTI, which consisted of 90,000 COVID-19-related tweets | BiLSTM,CNN, distilBERT,BERT,XLNET and ALBERT were applied | Accuracy 94.8% (BERT) |
| Sitaula et al. [28] | 2021 | The first work of sentiment analysis on Nepali COVID-19 tweets with three classes. They | Three different feature extraction methods—fastText-based feature extraction (ft), domain-specific probability-based | Accuracy 68.7% (Ensemble CNN) |
| | | prepared a public Nepali COVID-19 tweets dataset, called NepCOV19Tweets, for COVID-19-related sentiment analysis in the Nepali language | (ds), and domain-agnostic probability-based (da) feature extraction. Three different CNN models for the sentiment classification of tweets using three different feature extraction methods based on ft, ds, and da, respectively. In addition, for the results, an ensemble CNN model that captured the three different pieces of information was also designed. | |
| Shahi et al. [29] | 2022 | They used three different feature extraction methods—TF-IDF, FastText-based, and TF-IDF weighted FastText-based (hybrid) features—for the representation of COVID-19-related tweets written in the Nepali language. Here, the hybrid feature extraction in the Nepali language is a novel work in the study | They evaluated the performance of each feature extraction method on nine widely used machine learning classifiers | Accuracy 72.1% (SVM+RBF with hybrid feature extraction method) |
| Sitaula et al. [30] | 2022 | They designed a novel multi-channel convolutional neural network (MCNN), which ensembled the multiple CNNs, to capture multi-scale information for better classification. | Their proposed feature extraction method and the MCNN model were utilized for classifying tweets into three sentiment classes (positive, neutral, and negative) on the NepCOV19Tweets dataset | Accuracy 71.3% (MCNN) |
| Saadah et al. [31] | 2022 | Indonesian public opinion about COVID-19 vaccination tweets | BERT, IndoBERT, and CNN-LSTM were utilized for classifying COVID-19-related vaccination tweets | Accuracy 80% (IndoBERT) |

## The state State of the art of Sentiment Analysis of Tweets

In this section, we discuss some machine learning approaches, specifically the BERT model, to exploit COVID-19-related tweets.

BERT Model was utilized by Chintalapudi et al. [32] to explore the tweets by netizens from India during the COVID-19 lockdown. They labeled the text as joy, anger, sadness, and fear. Their model was compared with long-short term memory (LSTM) [33], support vector machines (SVM), and logistic regression (LR) models. They computed the accuracy of every sentiment separately. The BERT model outperformed the other models in terms of accuracy. A high prevalence of associated phrases and keywords among the tweets was recorded, some of them mentioned in Figure 4.

Müller et al. [34] proposed a transfer-oriented method dubbed COVID-Twitter-BERT (CT-BERT), pre-trained on the topic of COVID-19 with a large Twitter message corpus. The precise way to examine the performance of the domain-specific model was to utilize it on downstream tasks. 10-30% marginal enhancement was achieved by their model compared to the base model, BERT-LARGE. Their approach could be applied in different natural language processing tasks, including chatbots, question-answering, and classification.

COVID-19 fake news identification framework was presented by [35] based on ensembling learning and CT-BERT. They used additional data as a pre-processing step to achieve the weighted F1-score of 98.69. They empirically confirmed that BERT-based models achieved better binary classification outcomes.

In [36], a supervised machine learning technique was introduced to execute sentiment analysis of COVID-19 tweets. Tweepy library was exploited to extract tweets by a crawler. The pre-processing methods included cleaning the dataset, and the TextBlob library was employed to extract sentiments. Tweets were categorized as negative, positive, and neutral. LSTM architecture was also applied, but it yielded low accuracy as compared to other machine-learning classification techniques. Extra Trees Classifiers demonstrated superior performance than other models with 93% accuracy.

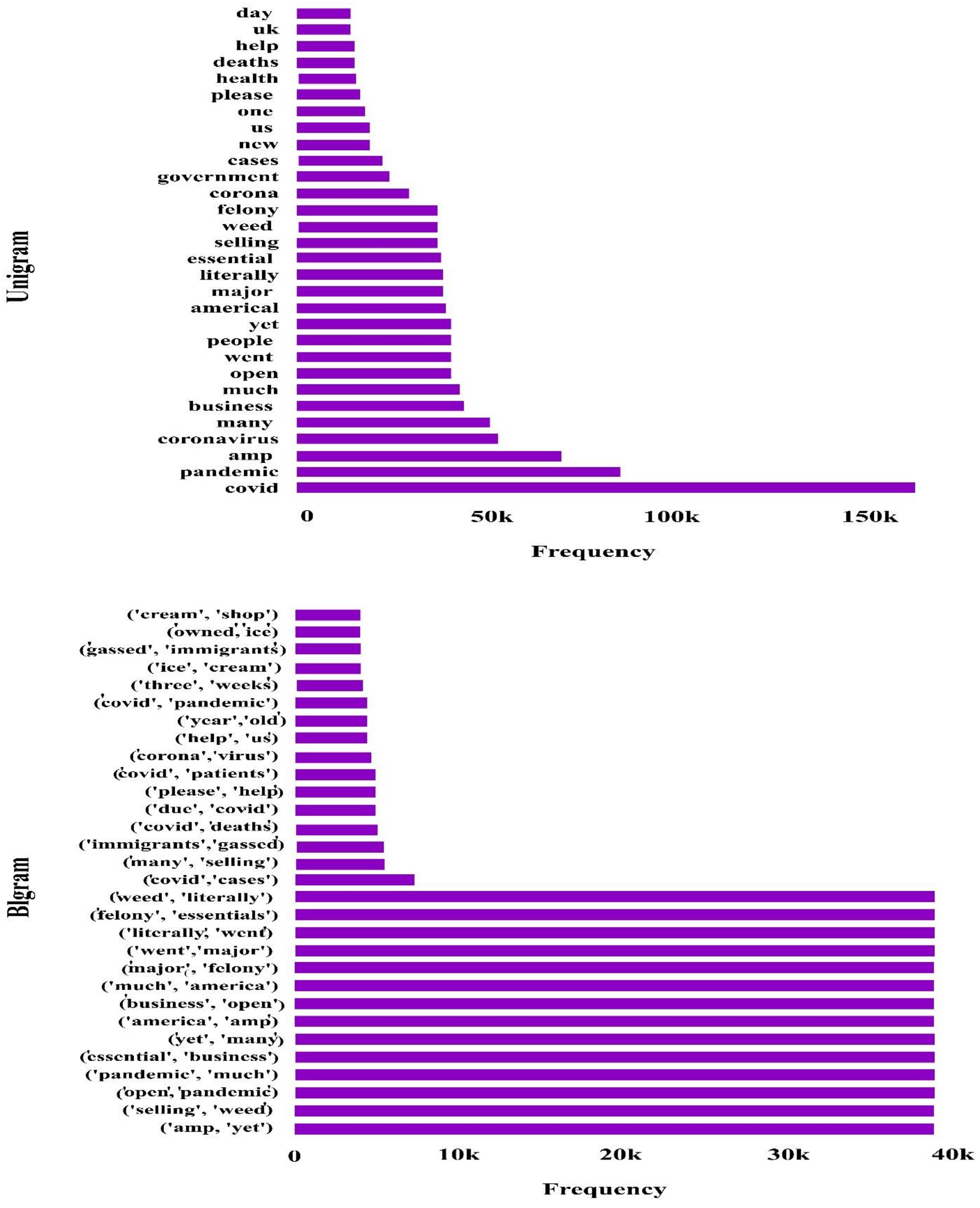

Figure 2. The counts of the most common uni-gram and bi-gram phrases as provided by the IEEE data port. These terms suggest that people discuss COVID-19-related government policies, lockdowns, and deaths [36].

Anti-vaccination tweet detection could provide vital information in devising strategies to mitigate such sentiments among different groups. The study [37] explored the anti-vaccination tweets during the COVID-19 pandemic using natural language processing techniques. They employed

BERT and Bi-LSTM with pre-trained GLoVe embeddings with Naïve Bayes (NB) and SVM. The BERT model showcased significant performance compared to other models.

Gencoglu et al. [38] utilized 26 million COVID-19-related tweets to devise a language-agnostic model based on BERT. For crisis management like the COVID-19 pandemic, quantifying the characteristics of public opinion is critical. Their study confirmed that lightweight classification techniques made comprehensive surveillance of public dialogue feasible by out-of-the-box exploitation of those demonstrations.

Chandra et al. [1] introduced a deep learning language model using LSTM for sentiment analysis of Indian people during the peak and rise of COVID-19 cases in India. Their architecture contained a state-of-art BERT language model and LSTM language model with a global vector embedding. Multi-label sentiment classification was applied with more than one sentiment expressed at once. They observed mostly optimistic tweets during the period, but a portion of the population expressed annoyance towards the authorities.

Pretrained language models were applied by Babu et al. [39] to gather information from tweets during the COVID-19 pandemic. They designed their framework as a binary text classification problem. Their CT-BERT model yielded an 88.7% F1-score, while their ensemble model containing SVM, RoBERTa, and CT-BERT showed an 88.52% F1-score.

Limiting negative emotions and irrelevant information is critical during catastrophic events like a pandemic. Malla et al. [40] presented a Majority Voting technique-based Ensemble Deep Learning (MVEDL) model to extract informative information from tweets. For training and testing, they used the “COVID-19 English labeled tweets” dataset. For the best performance with their model, they utilized state-of-the-art language models such as CT-BERT, BERTweet, and RoBERTa. The MVEDL model showed its efficacy by yielding high accuracy and F1-score compared to other deep learning counterparts.

Figure 3 [41] depicts the transformer family and clearly shows the relationship between the various transformers discussed in this paper, including BERT and RoBERTa.

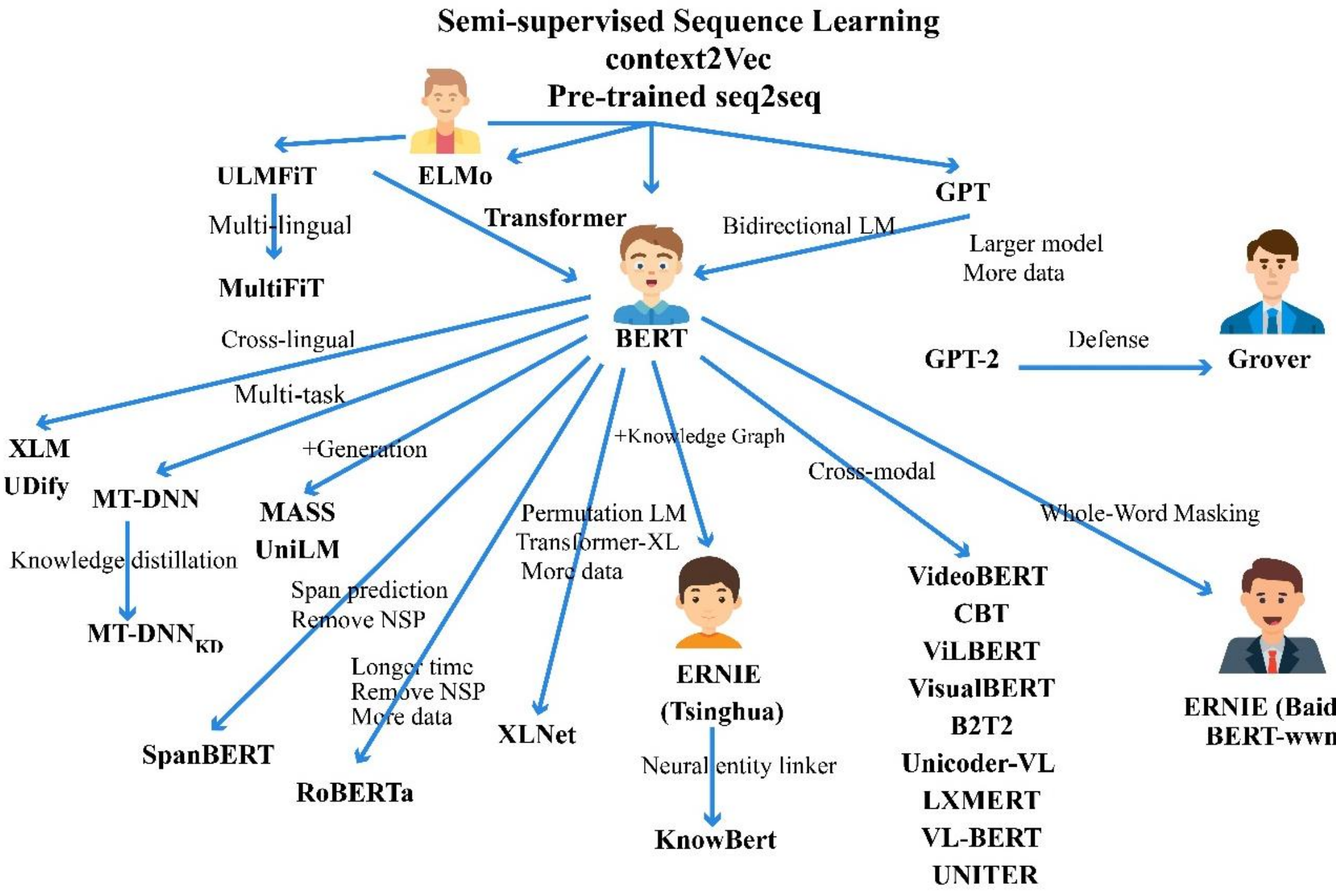

Figure 3. The Transformer family.

According to Figure 3, in transfer learning, the common practice is using a big generic model that has been pre-trained on a large-scale (text) dataset. The pre-trained model is later fine-tuned on a new possibly smaller dataset so that it will fit a particular task.

NLP covers various language processing applications such as sentiment analysis, chatbots, question-answering systems, machine translation, etc. The primary component of NLP approaches transformer models, which are used to predict the next character or word in a sequence. Examples of well-known transformer models are Google's BERT, Transformer-XL, XLNet, BiGRU, OpenAI's GPT-2, Transformer, RoBERTa, ALBERT, and ULMFiT.

1. ULMFiT (Universal Language Model Fine-Tuning), released in 2018, delivers good performance using NLP techniques. After training on the WIkitext 103 dataset, the model is fine-tuned for a new dataset. Based on ULMFiT, Anand et al. [42] proposed a suggestion-mining approach for forums and online reviews. They also presented a system description for SubTask A of SemEval 2019 Task 9. The objective is to determine whether a sentence contains a suggestion or not. For training the classification model, various pre-processing techniques are used. The trained model achieved an F1-score of 0.7011.

2. Transformer: this approach has a significant role in recent developments of NLP. In the past, Recurrent Neural Networks (RNNs) were used to implement machine translation and answering systems. The transformer approach outperforms RNNs and CNNs [43] thanks to a considerable reduction in training resource requirements. To model the relationship between words in a sentence regardless of their positions, a self-attention mechanism is used. Radford et al. [44] generated pre-trained models on different structured unlabeled texts. The models were then fine-tuned for specific tasks. Their approach was tested on 12 benchmark tasks. In the Stories Cloze Test for commonsense reasoning, RACE for question answering, and MultiNLI for text entailment, the performance was improved by 8.9%, 5.7%, and 1.5%, respectively.

3. OpenAI’s GPT-2: utilizing transformer models, the GPT-2 approach can predict the next word in a large volume of internet text (about 40GB). GPT-2 can generate high-quality conditional synthetic text samples outperforming rival methods on Wikipedia, news, or books. This approach has also performed well on tasks such as question answering, reading comprehension, summarizing, and translation from raw text. Without any direct supervision, Radford et al. [45] presented language models on WebText which consists of millions of web pages. GPT-2 achieved an F1-score of 55% on the CoQA dataset in question answering. The performance was further improved in a log-linear fashion.
4. BiGRU: The RNNs (Recurrent Neural Networks) are incapable of memorizing long sequences. This issue is addressed by Bidirectional Gated Recurrent Unit (BiGRU), which can predict a text paragraph with ease. This stems from the fact that BiGRU regularizes information flow using internal called gate mechanisms. BiGRU has already been used in various applications, such as speech recognition and synthesis. The fine-grained attention model [46] is based on a lengthy review in which the attention layer focuses on word-level, sentence-level, and paragraph-level features. This method has been validated on JD review and IMDB datasets.
5. Google’s BERT: Bidirectional Encoder Representations (BERT) considers both the left and right sides of a word to determine its context. BERT is capable of multitask-learning and, performing different NLP tasks simultaneously. BERT is the first bidirectional and deep system for unsupervised learning of NLP tasks. Biomedical text mining is considered very valuable among researchers. Lee et al. [47] investigated the possibility of adapting BERT for biomedical corpora, which resulted in the BioBERT (Bidirectional Encoder Representations from Transformers for Biomedical Text Mining) method. BioBERT outperformed BERT and other existing methods with 0.62% and 2.8% improvement on the F1-score for biomedical entity recognition and relation relaxation, respectively. BioBERT also achieved a 12.24% MRR improvement in biomedical question answering.
6. Transformer-XL: this method uses a new architecture for NLP without any fixed-length limitations [48]. Transformer-XL is significantly faster than Transformer. Using relative positional encoding is one of the critical ideas of Transformer-XL. In addition to single embedding, this method computes an embedding for each pair of tokens to capture the relation between them. While transformers can capture dependencies in longer sequences, they are limited to static-length scenarios. Dai Zihang et al. [49] proposed a method based on static length and fixed/constant temporal coherence. Their approach addressed the fragmentation problem and captured dependency for longer sequences.
7. XLNet: this method is a generalized auto-regressive model [50] capable of learning bidirectional contexts. XLNet can outperform BERT and other methods on several tasks by borrowing some of the techniques from BERT and Transformer-XL. The crux of XLNet is a novel objective function used during the pre-training phase. Integrating a two-stream attention mechanism and Transformer-XL, XLNet can achieve better performance in various tasks in machine vision and reinforcement learning. This method outperformed BERT in terms of error reduction on several datasets.
8. RoBERTa (Robustly optimized BERT approach) [51] was born by modifying the pre-training steps of BERT. Contrary to BERT, RoBERTa is trained on longer sequences, and NSP loss is not used. Moreover, during training, for feeding each series to the model, the generation of the masking pattern is done dynamically.

9. ALBERT [52] has been proposed as a lightweight version of BERT to deal with GPU/TPU memory limitations and the long training times of BERT. To this end, the number of parameters is reduced using cross-layer parameter sharing and factorized embedding parameterization without significant performance degradation. Despite having a lower number of parameters, ALBERT has more extensive architecture compared to BERT leading to computational complexity higher than BERTLARGE. However, ALBERT has outperformed BERTLARGE on GLUE, RACE, and SquAD benchmarks.

A study was conducted by Basiri et al. [53] to understand the general opinion of people in eight countries about the COVID-19 pandemic from their tweets. A hybrid fusion framework was proposed employing five deep-learning classification techniques. The prime findings were (i) every country had a unique sentiment pattern, (ii) the first reported infected case was correlated with a rise in information about the disease, and (iii) when there was a rise in active cases or deaths, negative sentiment was also at the peak.

Hayawi et al. [54] devised a new COVID-19 vaccine misinformation identification approach based on machine learning. They annotated and gathered COVID-19 vaccine-related tweets and utilized them to classify vaccine misinformation by training machine learning algorithms. They exploited the BERT model, LSTM, and XGBoost using 15,000 annotated tweets. BERT recoded an F1-score of 0.98 and hence proved its efficacy.

Another study by Vishwamitra et al. [55] analyzed hate speeches on Twitter against older adults and the Asian community triggered by the coronavirus pandemic. Their framework was trained to identify hate speech using the BERT strategy and applied the multi-headed attention mechanism of BERT to detect new keywords (100 keywords targeting older people and 186 keywords against the Asian community). Their study confirmed that BERT focused on particular word associations to identify hate speech in the Boomer-hate dataset, whereas BERT focused on varied attention between numerous words in the case of the Asian-hate dataset.

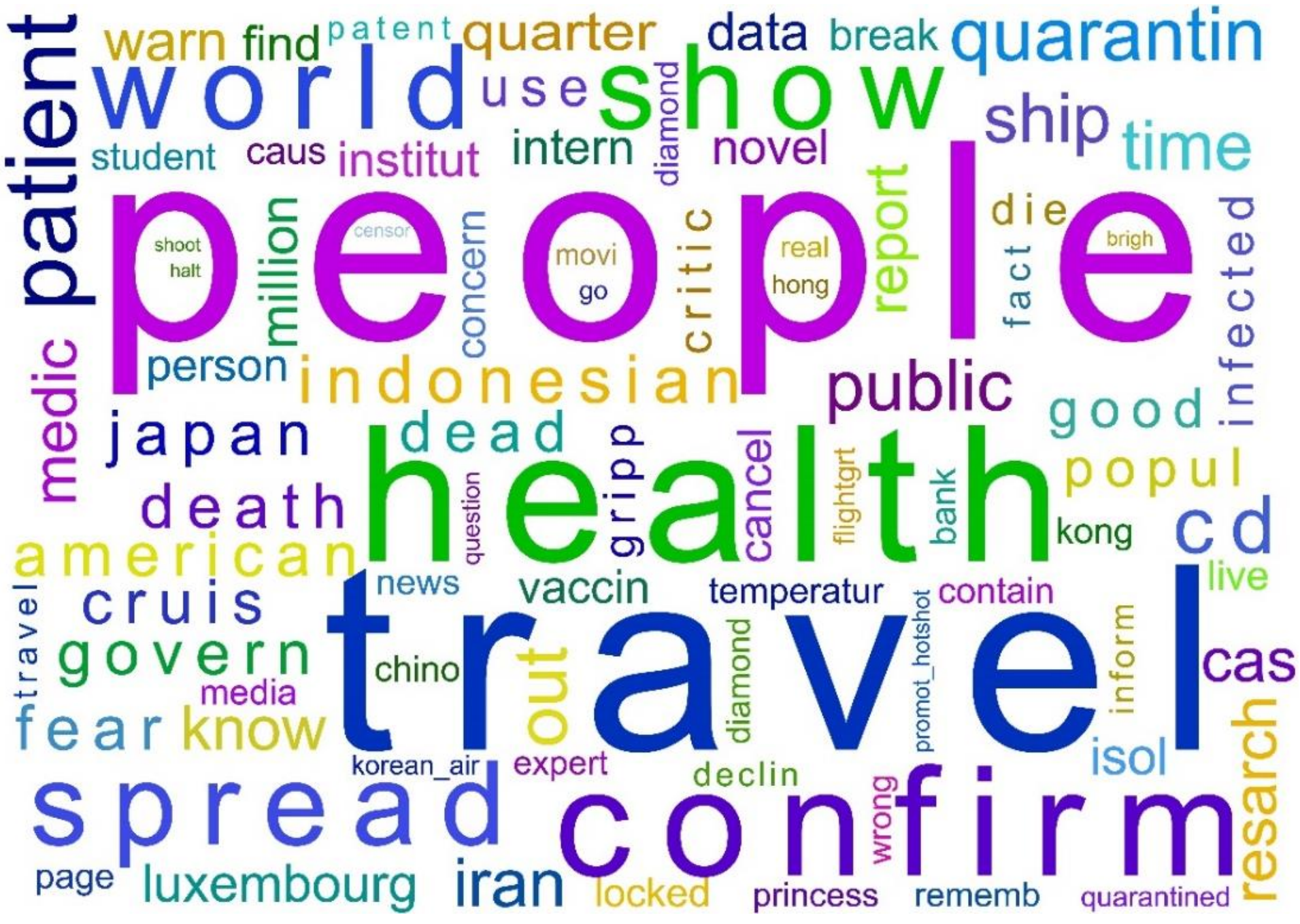

Figure 4. Word cloud of Luxembourg Tweets from 22 January 2020 to 1 March 2020 [56].

Two machine learning approaches were presented by Kabir et al. [57] for phrase extraction, and multi-label binary classification on an emotion dataset comprised of COVID-19 tweets applied for ten emotion labels classification and to choose a phrase that signified each emotion the best. They used a pre-trained technique, RoBERTa, with a custom Q&A head that tried to detect a word best suited for a particular feeling by taking the emotion label as a question. Their study showcased the adaption of the pandemic over time and staying more optimistic by people.

Valdes et al. [58] designed an automatic classification model for COVID-19-related Twitter posts. Their model was based on BERT. Their objective was to determine whether a model pre-trained on a corpus in the domain of interest could perform better than the one trained on a larger general domain corpus. They achieved encouraging F1-scores. They confirmed that a model trained with quality domain-specific data could achieve superior results compared to a model trained with a vast amount of general domain data.

A study by Tziafas et al. [59] proposed an ensemble system for misinformation identification in the context of the COVID-19 pandemic and called it the TOKOFOU system. It was based on six different pre-trained transformer-based encoders and fine-tuned each model on specific questions. The prediction scores were aggregated by applying the majority voting technique. The model achieved an F1-score of 89.7%.

Sadia et al. [60] explored Twitter opinions about the COVID-19 pandemic and devised a sentiment analysis model. They converted the Twitter data related to the pandemic into tokens. They utilized BERT and fine-tuned it with an extra classifier layer to categorize sentiments into three classes such as neutral, negative, and positive. Their model exhibited high scores in evaluation metrics.

The COVID-19 disinformation model was proposed by Song et al. [61] to classify pandemic-related misinformation that confused health-related issues among people. They presented a manually annotated COVID-19 misinformation corpus and a model for topic discovery and COVID-19 misinformation classification. They extensively examined misinformation classification concerning origin source, media type, wrong type, volume, and time.

The disinformation identification on Twitter can be categorized into two sub-tasks: (i) stance identification to detect whether the posts Disagree, Agree, or express no stance towards the wrong conceptions and (ii) extraction of disinformation being checked for integrity. Hossain et al. [62] gathered 6761 annotated tweets dataset called COVIDLIES1, for the COVID-19-related disinformation detection system. They evaluated their dataset on the prevailing NLP models and delivered the initial benchmarks to improve upon.

According to the studies, the Internet and mobile technologies are the crux of social media which lay the foundation for information dissemination, interactive communication, and content generation. Social media has claimed a pivotal role in the information ecosystem. In recent years, research on social media has attracted much attention from different domains. Social media analytics revolves around the development and evaluation of informatics tools and frameworks for the collection, monitoring, analyzing, summarizing, and visualization of social media data. As shown in Figure 5, social media analytics consists of three stages namely “capture,” “understand,” and “present” [3].

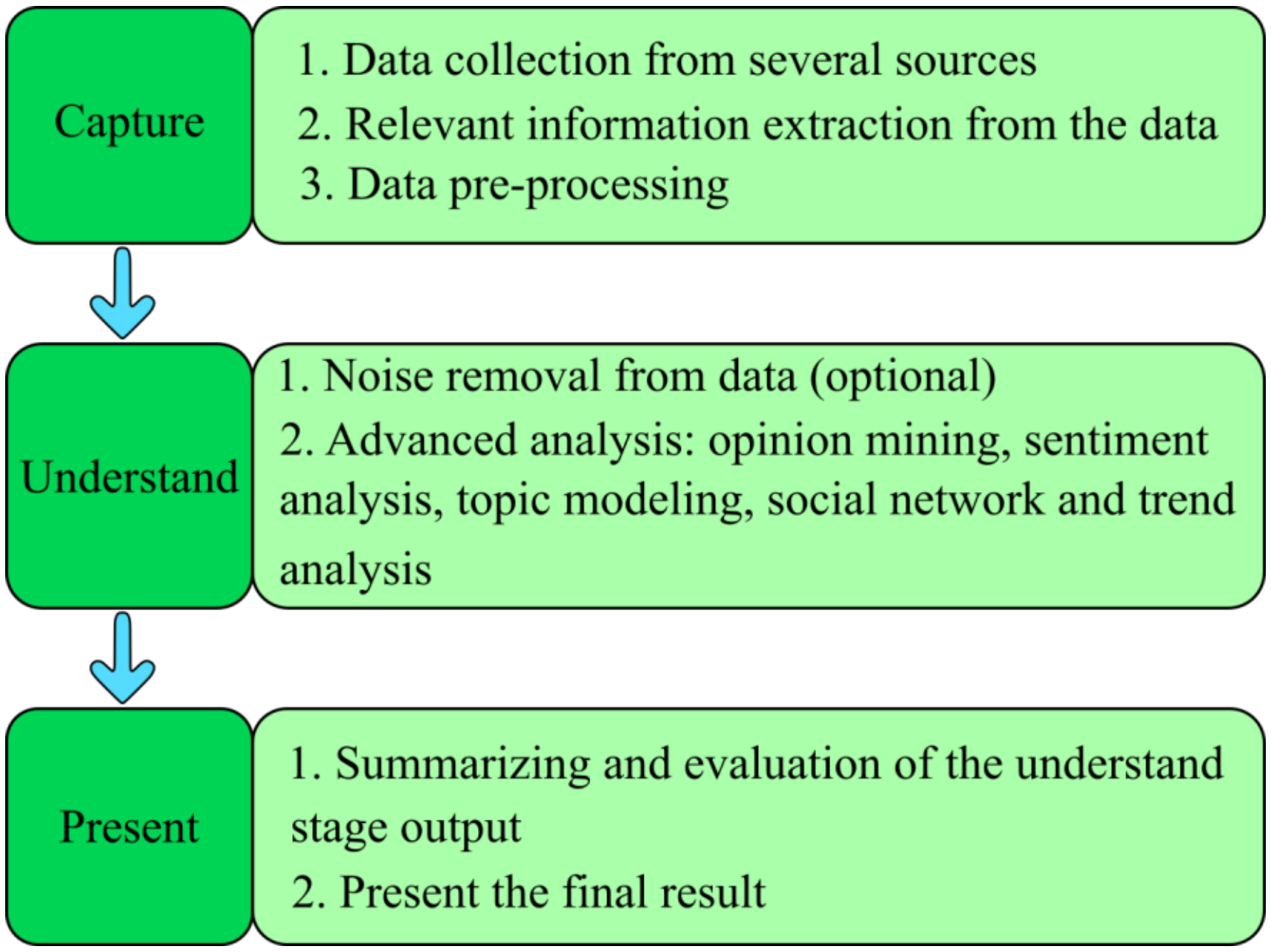

Figure 5. Process of social media analytics.

Based on Figure 5, we describe the steps in detail.

**Capture**: By gathering vast amounts of pertinent data from numerous social media sources, the capture stage helps find information on social media platforms connected to its activities and interests. These data are available and archived to be used to complete tasks. The preprocessed data are provided to the understanding stage through several preprocessing steps, such as data modeling, information and recorded linkage from diverse sources, stemming, part-of-speech tagging, feature extraction, and other syntactic and semantic procedures that enable the analysis.

**Understand**: The data acquired from various uses and sources during the capture stage typically include a significant percentage of noisy data that must be eliminated before helpful analysis. Then, various methods, from data mining, network analysis, machine translation, text, and natural language processing, can be used to retrieve insight from the purified data.

At this point, it is possible to generate a wide range of valuable data and trends about users, encompassing their histories, interests, problems, and social networks.

The comprehension phase, it should be noted, is at the center of the whole social media evaluation. The information and analytics at this time will be significantly impacted by its results, which will greatly assist businesses in making decisions.

**Present**: The findings of various analytics are analyzed, compiled, and presented understandably as the final step. Helpful information can be explained using a variety of visualization methods.

## Discussion

Social media are an integral part of today's life, and they have too many users all over the world. For example, Facebook, Instagram, and Twitter have 2910, 1487, and 436 million users as of January 2022, respectively [63]. Hence, these platforms provide a vast inexhaustible resource of data, to exploit in different research topics.

On the other hand, such a large amount of information has made it impossible for humans to analyze and review this data, and after that, the broad demand for data analysis in recent years has led to a growth in the quest for automatic systems that can categorize the sentiment analysis which is the sentiment assigned to a sentence.

Among the topics that have been discussed and trended ubiquitous in social media over the last two years was the COVID-19 pandemic. Coronavirus disease 2019, or in brief, COVID-19, is a contagious disease caused by the SARS-CoV-2 virus. The first recognized case was detected in Wuhan, China, in December 2019 [64]. The disease spread worldwide so fast, resulting in the COVID-19 pandemic that affected the lives of people worldwide in many aspects. Consequently, as of March 2020, over 628 million tweets were sent or retweeted [65]. These facts clearly show the importance and complexity of analyzing tweets related to COVID-19.

In this research study, we reviewed the state-of-the-art BERT and deep CNN models for sentiment analysis of COVID-19 tweets. It is worth noting that recently, CNN and also, Recurrent Neural Networks (RNN)-based models are ubiquitous algorithms for sentiment analysis and have been widely implemented in this field of study. However, relying merely on the CNN-based model in sentiment analysis may lead to an extremely low accuracy which is a result of disregarding semantic associations that exist within the context of the review texts. Therefore, a suitable and efficient alternative for CNN-based models is to use BERT models. BERT implements masked language methods to empower pre-trained deep bidirectional language model representations, which decrease the requirement for a wide range of massively-engineered context-specific frameworks. Moreover, it is the first finetuning-based representation method that obtains state-of-the-art implementation on many token and sentence-level tasks, which dominates and overcomes various task-specific frameworks.

In the conclusion of this section, the performance of several literature studies on sentiment analysis of COVID-19 tweets using BERT or CNN-based models is shown in Table 2.

**Table 2.** The performance of several literature studies on sentiment analysis of COVID-19 tweets.

| **Research Study** | **Method** | **Dataset** | **Number of Samples** | **Accuracy (%)** | **Precision (%)** | **Recall (%)** | **F1-Score (%)** |
|---|---|---|---|---|---|---|---|
| Babu et al. [39] (2020) | Ensemble of CT-BERT, RoBERTa, and SVM | WNUT-2020 Task 2 | 10,000 | n/a | 89.24 | 87.82 | 88.52 |
| Gencoglu [38] (2020) | SVM + LaBSE embedding | Publicly available COVID-19 tweets between 4 January and 5 Apr (2020) | 26,759,164 | 86.92 | n/a | n/a | 88.10 |
| Rustam et al. [36] (2021) | Extra Tree Classifier using concatenated features | IEEE data port (on May 31, 2020) | 7528 | 93 | 90 | 89 | 89 |
| Basiri et al. [53] (2021) | Ensemble deep Learning using five base | Stanford Sentiment140 | 1,600,000 | 85.8 | n/a | n/a | 85.8 |
| | Learners | | | | | | |
| To et al. [37] (2021) | BERT | Anti-vaccination tweets between 1 January and 23 August (2020) | 1,651,687 | 91.6 | 93.4 | 97.6 | 95.5 |
| Malla & Alphonse [40] (2021) | Majority Voting-based Ensemble Deep Learning(MVEDL) | WNUT-2020 Task 2 | 10,000 | 91.75 | 89.94 | 93.58 | 91.14 |
| Glazkova et al. [35] (2021) | g2tmn | COVID-19 Fake News Dataset | 10,700 | n/a | n/a | n/a | 98.69 |
| Hayawi et al. [54] (2022) | BERT | ANTi-Vax | 15,073 | 98 | 97 | 98 | 98 |

The research projects we employed for our study have the limitation of ignoring the effects of worldwide COVID-19 news and data on the general sentiment of some other countries. Also, there are two types of sentiment classification. Positive and negative sentiment is categorized as binary. However, text sentiment can be separated into more than two classifications as multi-class sentiment. As an example, consider three classes whose sentiment is categorized as positive, negative, or neutral. Some of the research projects here include a classification of news headlines into either positive or negative classes as their primary objective (binary). However, there is unquestionably a chance that multi-class sentiment classification may lead to more effective outcomes.

The computational resources required to train, fine-tune, and derive conclusions are the fundamental limitations of employing BERT and other large neural language models. However, more recent studies have suggested many solutions to deal with this problem [66].

Finally, one of the significant limitations in sentiment analysis of messages sent about a specific phenomenon in a social network with many users from all over the world is that moods, opinions, and attitudes toward that topic may be influenced by culture. Different communities' views and beliefs are not well reflected in the messages, resulting in errors in the analysis performance. Furthermore, in a phenomenon such as the COVID-19 pandemic, fake news or incorrect scientific results can lead to the sending of mass messages on social networks that do not reflect reality, and such challenges are aimed at sentiment analysis.

## Conclusions and Future research directions

Since tweets are written in informal language, they use misspelled words and employ careless grammar. Sentiment analysis involves various Natural Language Processing (NLP) [67, 68] tasks, including sarcasm detection, aspect extraction, and subjectivity detection. Accordingly, deep convolutional neural network architectures are capable of recognizing binary sentiment in Twitter data, but in comparison with other sentiment classification tasks, they fail to perform nearly as well [69, 70]. With pandemic diseases like COVID-19, which require many resources and equipment types, deep learning models perform better with a more extensive and comprehensive dataset [71, 72]. Therefore, it is necessary to conduct further research to generate and share a complete dataset for research purposes. This paper determined ways to improve future research. Most of the research is limited to English-language text, which was considered a selection criterion. Therefore, the results do not reflect comments made in other languages, such as Persian language or another one. In addition, much of the research was limited to the remarks retrieved in short timelines. Consequently, the period between the completion of the research and the subsequent studies may have affected the timeliness of the results. So, the other directory should be related to different languages' text as a selection criterion. A new sentiment analysis model using BERT and deep CNN can be developed by researchers. Based on a sentiment dictionary inserted into the word vector of some models, recurrent neural networks can extract forward and reverse contextual information [73]. Therefore, BiLSTM might be able to emphasize different words in a text more or less by adding an attention mechanism to its output. There is also the possibility of using a high-quality and lightweight BERT model to perform downstream tasks.

## Acknowledgements

Not applicable.

## Funding

This study received no funding.

## Author Contributions

JHJ designed the study. Literature search performed by SH. Figures production done by MAN and RA. The conclusion and abstract have been written by RB. FF and SH have written the limitations and contributions of the study. The final version of the paper has been edited by JHJ, SH, MAN, RB, RA and AT. RB supervised the project, and RL co-supervised the study. All authors have read and approved the final manuscript.

## Conflict of Interest

The authors declare that the research was conducted without any commercial or financial relationships that could be construed as a potential conflict of interest.